\documentclass[letterpaper, 10 pt, conference]{ieeeconf}  

\IEEEoverridecommandlockouts                              

\usepackage{caption}
\usepackage[skins]{tcolorbox}  
\usepackage{subcaption}
\usepackage{multicol}

\usepackage{mathtools}
\usepackage{graphics} 
\usepackage{epsfig} 
\usepackage{mathptmx} 
\usepackage{times} 
\usepackage{amsmath} 
\usepackage{amssymb}  
\usepackage{graphicx}
\usepackage{epstopdf}
\usepackage{pdfpages}
\usepackage{cite}
\usepackage{amsbsy}
\usepackage{bm}
\usepackage{url}
\usepackage{accents}

\pdfoutput=1
\usepackage{hyperref}
\hypersetup{
    colorlinks=true,
    linkcolor=black,
    citecolor=black,
    filecolor=black,
    urlcolor=black,
}

\def\horizontaldistance{\kern2pt}

\title{\LARGE \bf
Online Bipedal Locomotion Adaptation for Stepping on Obstacles Using a Novel Foot Sensor
}

\author{Pezhman Abdolahnezhad$^{1}$, Aghil Yousefi-Koma$^{1}$, Amirhosein Vedadi$^{1}$,  Kasra Sinaei$^{1}$\\
Behnam Maleki$^{1}$ and  Milad Shafiee$^{2}$
\thanks{$^{1}$Center of Advanced Systems and Technologies (CAST) School of
Mechanical Engineering, College of Engineering, University of Tehran,
Tehran, Iran.
        {\tt\small aykoma@ut.ac.ir}}%
      \thanks{$^{2}$École polytechnique fédérale de Lausanne (EPFL) 
        }%
       }

\makeatletter
\newcommand*{\rom}[1]{\expandafter\@slowromancap\romannumeral #1@}
\makeatother
\begin{document}

\maketitle
\thispagestyle{empty}
\pagestyle{empty}

\begin{abstract}
In this paper, we present a novel control architecture for the online adaptation of bipedal locomotion on inclined obstacles. In particular, we introduce a novel, cost-effective, and versatile foot sensor to detect the proximity of the robot's feet to the ground (bump sensor). By employing this sensor, feedback controllers are implemented to reduce the impact forces during the transition of the swing to stance phase or steeping on inclined unseen obstacles. Compared to conventional sensors based on contact reaction force, this sensor detects the distance to the ground or obstacles before the foot touches the obstacle and therefore provides predictive information to anticipate the obstacles. The controller of the proposed bump sensor interacts with another admittance controller to adjust leg length. The walking experiments show successful locomotion on the unseen inclined obstacle without reducing the locomotion speed with a slope angle of $12^\circ$. Foot position error causes a hard impact with the ground as a consequence of accumulative error caused by links and connections' deflection (which is manufactured by university tools). The proposed framework drastically reduces the feet' impact with the ground.

\end{abstract}

\section{INTRODUCTION}
One of the significant advantages of legged robots over other types of mobile robots, such as wheeled robots, is that they possess the capability of traversing uneven terrains and functioning in harsh environments \cite{Humanoid2014,wahrmann2017modifying}. Stabilizing bipedal robots' gait while operating in an unknown environment, specifically when they walk on an uneven surface, is crucial for achieving robust bipedal mobilization. Besides the unknown profile of the ground, uncertainty, clearance, and errors in actuators, links and joints are other sources of instability that can mitigate the robot’s body fluctuations even when walking on perfectly flat surfaces. Another major challenge regarding the smooth walking of position-controlled robots is the disturbance exerted on the robot body by contact force impacts from each sole at the beginning of the support phase \cite{kajita2005running}. This problem becomes more significant when the walking pace of the robot increases, so minimizing the effects of ground reaction forces impact that occurs in the swing sole contact change is essential for reaching higher walking speeds. Humanoids' reliable, human-level walking ability depends on maintaining a smooth and continuous contact change between the robot's soles and the ground with minimal contact impact \cite{xu2010force}. The most utilized and venerated criteria for checking the stability of a bipedal robot is Zero Moment Point (ZMP); this is a necessary condition. Dynamic stability of walking is not within the scope of this work; however, it is worth mentioning that the ZMP stability criterion is a necessary but not sufficient condition for contact wrench stability \cite{hirukawa2006universal}. As a result of that, when dealing with position-controlled humanoid robots, the individual contact force distributed at each sole should be studied to address the sole impact disturbance problems \cite{caron2015stability}. Admittance controllers are used to control the forces, reduced model\cite{ramuzat2021comparison,zhou2016stabilization,li2012passivity} or torques of the position-controlled humanoid robots in a effector (wrist or ankle) \cite{pfeiffer2017nut,keemink2018admittance}.

\begin{figure}[]
\centering
\includegraphics[scale=0.63, trim ={4.0cm 15.1cm 2.0cm 2.0cm},clip]{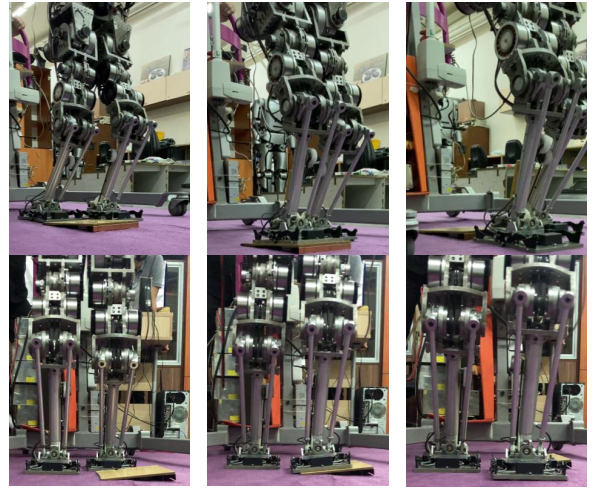}
\caption{Snapshots of Surena V  walking on inclined obstacles}
      \label{fig:surena4}
\end{figure}

Such controllers have been exclusively investigated in recent years. Kajita et al. designed a framework and analyzed its performance on the HRP4-C robot that could stabilize the robot that makes slight modifications to the preplanned walking patterns \cite{kajita2010biped}. Joe et al. demonstrated a posture controller that adjusts the position and orientation of ankles based on the sensory data that could be obtained from F/T sensors; the proposed framework allowed the robot to perfectly adapt its ankle to the unknown ground it is traversing \cite{joe2019robust}. The ZMP distributor and force distributor logic used in Joe's work follows similar logic to Kajita's strategy of finding the desired wrench of the planned walking pattern \cite{joe2019robust}. Caron et al. utilized another whole-body admittance controller to stabilize the stair climbing of the HRP4 robot by exploiting the foot wrench measured data and estimated DCM (divergent component of motion) and ZMP positions \cite{caron2019stair}. 

The aforementioned control strategy does not require the robot to replan its footsteps. As a result, generating a new set of trajectories for the CoM (Center of Mass) and the ankles is not necessitated. Since rescheduling gait patterns is computationally demanding and difficult to implement in real-time cases, this could be highly beneficial for algorithms that rely on admittance controllers.

Sygulla et al. introduced a hierarchical control strategy to overcome the blind walking of bipeds on uneven surfaces. The control strategy has been tested on Lola humanoid robot. The early contact detection allows their robot to reduce contact forces on impact whenever the swing leg faces an unprecedented obstacle. The proposed control scheme by Sygulla relies on the F/T sensor data and force pad switches mounted to the soles of the robot. Force pad switches could provide essential data for detecting early contact \cite{sygulla2020force}. Despite the acceptable performance of Sygulla's control scheme, relying on their sensors, the robot can only sense early contacts after the sole comes into contact with the ground and touches the surface. These force pad switches can't determine the distance to the ground at different walking speeds. Employing proximity push sensors which we will introduce in this paper, could better predict and absorb impacts from contact force by knowing the distance to the ground. These types of sensors have already been introduced to walk on uneven surfaces \cite{hashimoto2005development, kang2010realization, yousefi2021surenaiv}. Specifically Kang et al. have designed a foot mechanism for the WABIAN-2R robot consisting of four spikes, each of which has an optical sensor to detect ground height. They developed a simple P controller with the feedback of these sensors.





This paper presents a control scheme to assist the lower limb of the bipedal robot Surena V in minimizing the impact caused by contact forces in swing-support change. This set of controllers is essential for walking at higher speeds. The proposed framework also addresses walking stabilization over uneven surfaces. Several admittance controllers were employed simultaneously to minimize the fluctuations and disturbances caused by rugged terrains. These controllers utilize the feedback from F/T sensors mounted to the bottom of ankle links and four spring-loaded push sensors mounted to the four corners of each sole. The push sensors enable the robot to better estimate the obstacles and the surface on which it is walking. Adjustments made by the abovementioned admittance controllers to the planned trajectories of ankle position and orientation are essential for increasing the robots walking pace and gait robustness.

The framework comprises an admittance controller, which is in charge of manipulating the height of the ankles; this is done by adding its generated control output to the preplanned trajectories for the ankles. The feedback of this controller comes from the force component of a couple of three-axis F/T sensors mounted at the bottom of each sole. Another controller is employed to boost the ankle height admittance controller's performance by detecting the swing ankle's early contact and adding a controlled output to its preplanned position based on the data received from push (bump) sensors. The data from sole bump sensors could also be utilized for altering the orientation of the sole to adapt the support leg to its adjacent ground. 

The lower limb of the Surena V humanoid robot was used for experimental tests of the framework. Surena humanoid robot has 12 Degrees of Freedom (DoF) in its lower limb (6 DoFs in each leg). Based on the experiments, it has been concluded that each controller, ankle height admittance controller, and push sensor controller, increased the stability of the gait; however, the optimal performance was achieved when the combined output of all three controllers was used to modify the preplanned ankle trajectories.

This paper is structured as follows; in the next section, theories related to walking pattern generation, admittance controllers, and detailed control schemes are elaborated. Then in the third section, our experimental setup and results from executing the mentioned control scheme will be presented, and its performance will be discussed. Finally, in the fourth section, the discussions are concluded.
\section{Architechture}

\begin{figure}[]
\centering
\includegraphics[scale=0.45, trim ={0.0cm 0.0cm 0.0cm 0.0cm},clip]{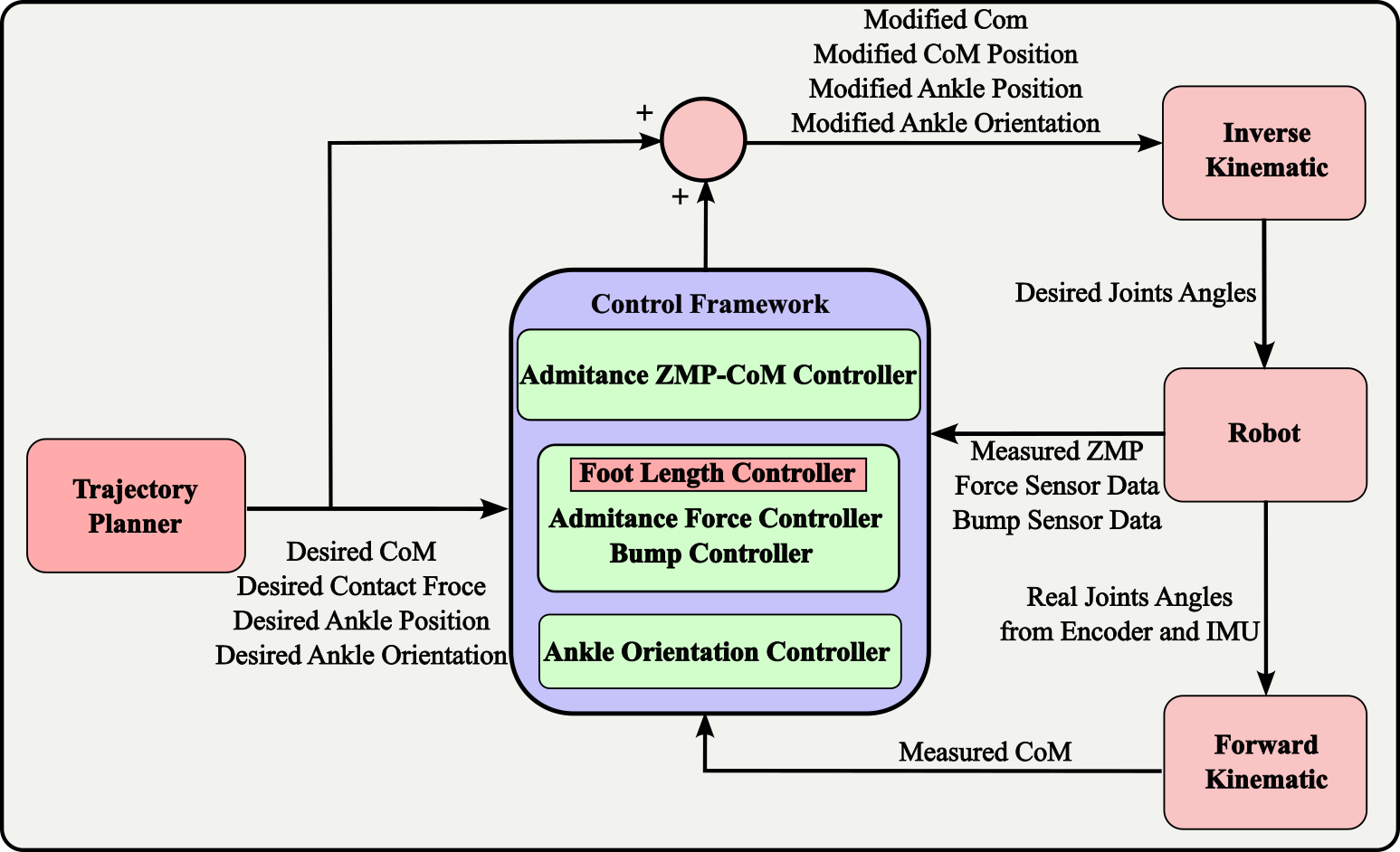}
\caption{Control Architecture of Online Adaptation}
      \label{fig:diagram}
\end{figure}
\begin{figure}[]
\centering
\includegraphics[scale=0.45, trim ={1.0cm 19.7cm 0.0cm 1.0cm},clip]{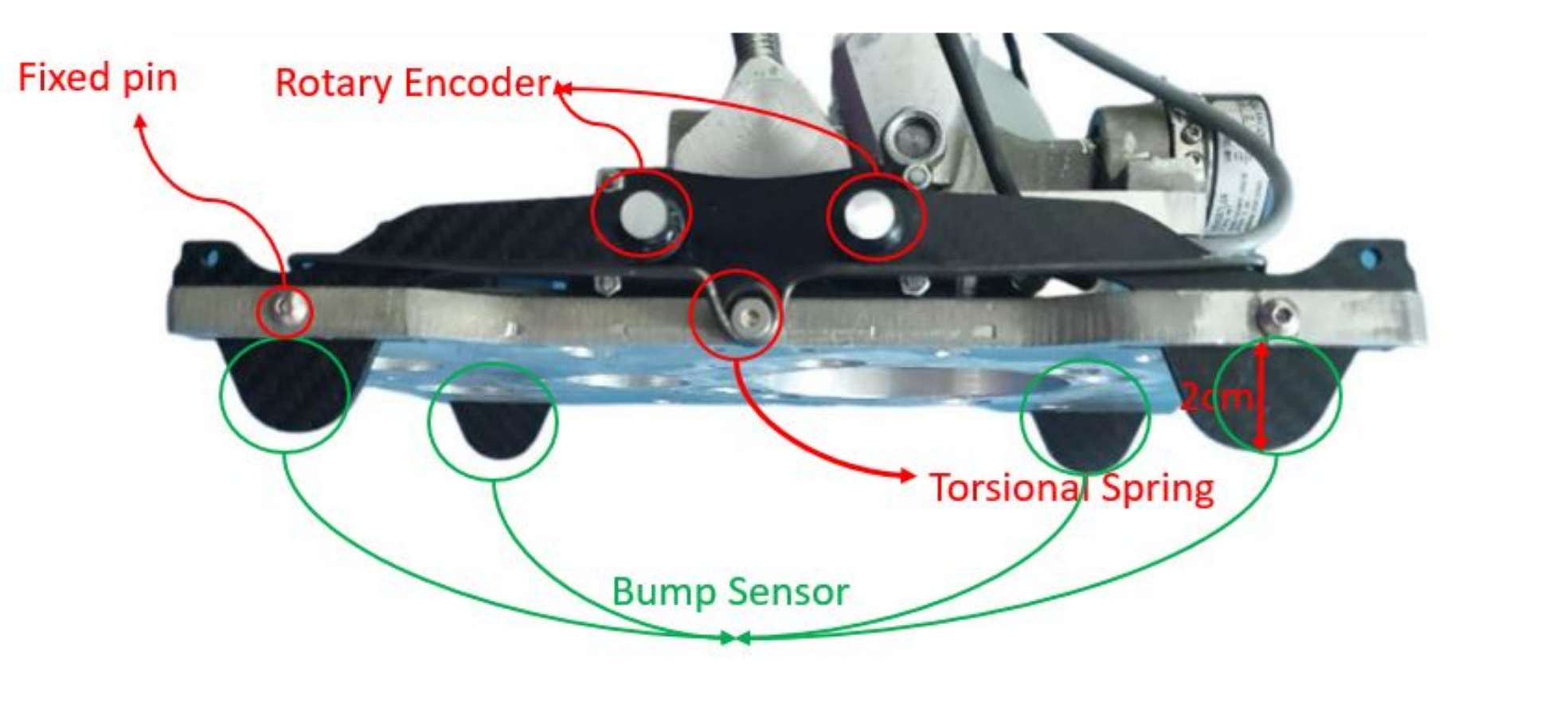}
\caption{Bump sensors used in the robot}
      \label{fig:Bumpsensors}
\end{figure}
As illustrated in Figure 2, the control architecture of our robot comprises two main parts: The trajectory planner and the control framework. The trajectory planner generates the trajectory of the desired  center of mass (CoM), zero moment point (ZMP), and ankle positions in real time based on the algorithm presented in \cite{englsberger2015three}. Trajectory  parameters have been extracted using non-dominated sorting genetic algorithm II (NSGA-II) optimization to plan a trajectory that minimizes energy consumption and enhances stability according to \cite{vedadi2021bipedal}. The control scheme includes the introduction of a new sensor so-called bump sensor, and three controllers: Foot Length, Ankle Orientation, and Admittance ZMP-COM controller, which will be elaborated on in the incoming sections.

\subsection{Bump Sensor}
Four rotary contact sensors are used on the four corners of each foot, as described in Figure 3. Each of these sensors is bounded with a torsional spring which it's torsional constant is about $880 \frac{kN.m}{rad}$. Also, fixed pins are used to prevent bump sensors from moving laterally. It compresses to 2 cm when the robot's foot is about to make contact with the ground and returns to its initial position when the foot is in the swing phase. One of the purposes of using this new sensor is ankle adaptation. Another goal is to calculate the actual height of the robot's feet from the ground in order to avoid hitting the ground. Compared to the F/T sensor, the design complexity of the bump sensor is significantly lower. As a result, bump sensors could be fabricated and utilized conveniently at a much lower cost.

\begin{figure}[]
\centering
\includegraphics[scale=0.7, trim ={1.5cm 15.0cm 7.0cm 8.0cm},clip]{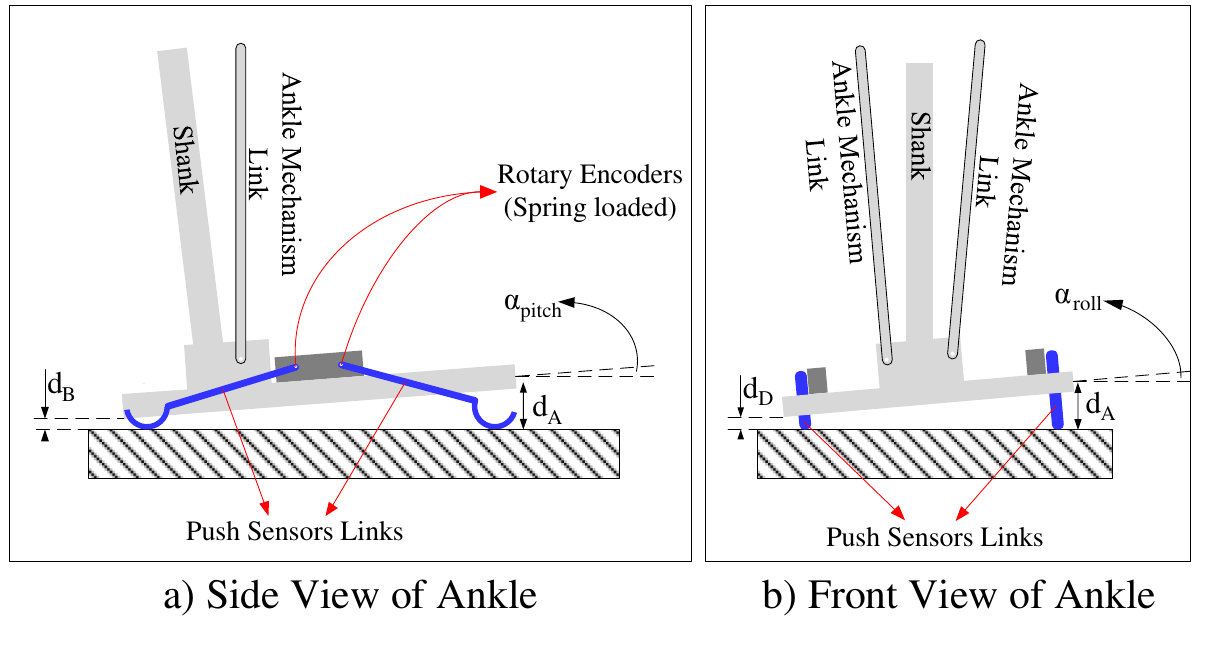}
\caption{Schematic diagram of foot sensors and their configuration in
the ankle link}
      \label{fig:bumpsensor}
\end{figure}

\subsection{Foot Length Controller}
The foot length controller consists of two layers: the first one is to control the change of the robot's leg length by converting contact force into the ankle's vertical displacement during walking. For this admittance controller, the desired forces ($F_{z,d}^L$,$F_{z,d}^R$) are obtained by distributing the force by using the desired ZMP, according to the method mentioned in  \cite{kajita2010biped}:
\begin{equation}
e= (F_{z,d}^L-F_{z,d}^R)-(F_m^L-F_m^R) 
\label{eq:eq1}
\end{equation}

\begin{equation}
\Delta \dot z = k_p.e-k_r.\Delta z
\label{eq:eq2}
\end{equation}


\begin{figure}[]
\centering
\includegraphics[scale=0.5, trim ={3.2cm 18.0cm 0.0cm 2.5cm}]{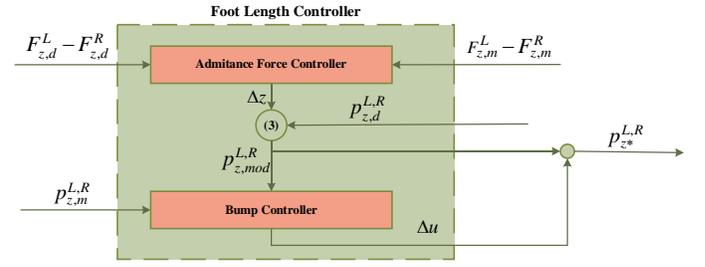}
\caption{Foot Length Controller}
      \label{fig:FootLengthController}
\end{figure}
Where $k_p$ is a proportional feedback gain, $k_r$ is a gain to retrieve neutral points, and $F_m^L$,$F_m^R$ are the real force values of the robot's soles, which are measured by force/torque sensors. Using  (2), the value of $\Delta z$ is adjusted to reach the target $(F_{z,d}^L-F_{z,d}^R)$. The relative height changes  are applied to the ankle height as follows \cite{kajita2010biped}:
\begin{equation}
p_{z,mod}^L = p_{z,d}^L- 0.5 \Delta z , \hspace{0.1cm}  p_{z,mod}^R = p_{z,d}^R + 0.5 \Delta z
\label{eq:eq3}
\end{equation}
As mentioned in the conclusion of this article \cite{caron2019stair}, the effects of swing foot touch down impact could not be eliminated solely by the first controller; to address this cause another controller should accompany the first controller. This is the reason for using the second layer in this part. Unlike force sensors, bump sensors perceive incoming contacts before feet reach the ground. By changing the length of the robot's legs, this controller ameliorates the intensity of the touchdown impact swing phase. The actual height of the robot's feet ($p_{z,m}^L$,$p_{z,m}^R$) is measured according to the average value of the bump sensors. The desired value of the controller input is the same as the output of the Admittance Force Controller (3), that is ($p_{z,mod}^L$,$p_{z,mod}^R$):

\begin{equation}
p_{z,m}^{L,R} \propto d_{avg}^{L,R} = \frac{d_{A}+d_{B}+d_{C}+d_{D}}{4}
\label{eq:eq4}
\end{equation}

\begin{equation}
e^{L,R}= p_{z,m}^{L,R} - p_{z,mod}^{L,R}
\label{eq:eq45}
\end{equation}

\begin{equation}
\Delta \dot u^{L,R}=k_p e^{L,R}- k_r \Delta u^{L,R}
\label{eq:eq6}
\end{equation}

This controller (6) works for the left and right foot in a certain range of single support phase and when the distance between the foot and the ground is less than 2 cm. The relative height changes $\Delta u$ are applied to the ankle height as follows:

\begin{equation}
p_{z*}^L = p_{z,mod}^L+ \Delta u^L, \hspace{0.1cm}  p_{z*}^R =  p_{z,mod}^L + \Delta u^R
\label{eq:eq7}
\end{equation}

Figure 5 shows the operation of foot length controller.

\begin{figure}[]
\centering
\includegraphics[scale=0.6, trim ={0.75cm 1.0cm 0.3cm 0.2cm},clip]{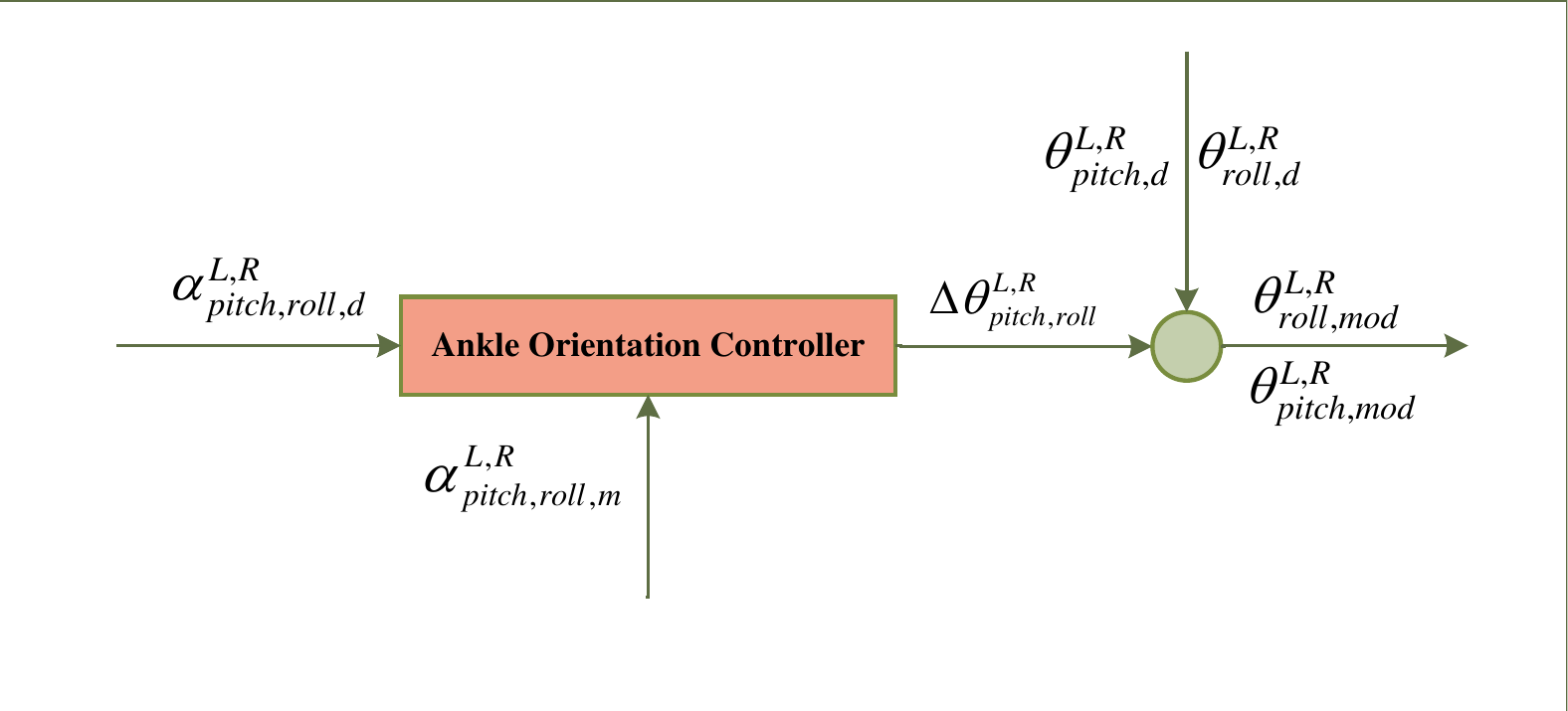}
\caption{Ankle Orientation Controller}
      \label{fig:FootLength}
\end{figure}

\subsection{Ankle Orientation Controller}
According to Figure 4, the measured pitch and roll angles between the robot’s sole and ground ($\alpha_{pitch,m}$,$\alpha_{roll,m}$),  are proportional to the values of four bump sensors as follows:
\begin{equation}
\alpha_{pitch,m} \propto d_{B}+d_{C}-d_{A}-d_{D}
\label{eq:eq8}
\end{equation}

\begin{equation}
\alpha_{pitch,m} \propto d_{A}+d_{D}-d_{C}-d_{D}
\label{eq:eq9}
\end{equation}

The ankle orientation controller adapts the sole of the robot's foot to the ground surface.
\begin{equation}
e_{pitch,roll}^{L,R} = \alpha_{pitch,roll,m}^{L,R} - \alpha_{pitch,roll,d}^{L,R} 
\label{eq:eq10}
\end{equation}

\begin{equation}
\Delta \dot \theta_{pitch,roll}^{L,R} = k_p \alpha_{pitch,roll,m}^{L,R} - k_r \Delta \theta_{pitch,roll}^{L,R}
\label{eq:eq11}
\end{equation}

$\alpha_d$is the angle we want the sole of the foot to have with the ground. When the robot walks, the sole of the foot should be parallel to the ground, so its value is set to zero. The output of the controller, $\Delta \theta_{pitch,roll}^{L,R}$ is finally added with the desired pitch and roll values,  $\theta_{pitch,d}^{L,R}$, $\theta_{roll,d}^{L,R}$ of the ankle.

\begin{equation}
\theta_{pitch,mod}^{L,R} =  \theta_{pitch,d}^{L,R} +  \Delta \theta_{pitch}^{L,R} \hspace{0.1cm} , \theta_{roll,mod}^{L,R} = \theta_{roll,d}^{L,R} +  \Delta \theta_{roll}^{L,R}
\label{eq:eq12}
\end{equation}

Figure 6 shows the operation of ankle orientation controller.
\subsection{Admittance ZMP-CoM Controller}
In third layer of the control framework, the desired CoM trajectory modifies according to the ZMP deviations from its reference trajectory. The stability of this controller has been proven in \cite{choi2007posture}. 
\begin{equation}
e_{zmp} =  r_{d}^{zmp}  -r_{m}^{zmp} , \hspace{0.1cm} e_{com}=x_d^{com}-x_m^{com}
\label{eq:eq13}
\end{equation}

\begin{equation}
u =  -k_p.e_{zmp} + k_c.e_{com} 
\label{eq:eq14}
\end{equation}

Where the $ r_{d}^{zmp}$ and  $X_{d}^{com}$ are the desired positions from trajectory planner, $ r_{d}^{zmp}$ and  $X_{d}^{com}$ are the actual positions of CoM and ZMP measured from real bipedal robot, respectively. The output of the ZMP-CoM Controller (14) is the amount of changes in the velocity of the center of mass, which is finally added to the com position after integration.
\begin{equation}
X_{mod}^{com} = \int u dt +  X_{d}^{com}
\label{eq:eq15}
\end{equation}
\section{Experiments and Results}
We implemented the discussed controllers on the lower limb of the Surena V robot and released the source code of it as an open-source project \cite{walkingController}. The loop rate of the implemented controller is 200 Hz, and it runs on an Intel Core i7 processor, which works with a frequency of 3.6 GHz and 16 gigabytes of RAM. We used Robot Operating System (ROS) to share data between the robot and the main processor. To show the performance of the added sensors and controllers and their effect on walking, we investigated the robot in high-speed walking and walking over unforeseen inclined obstacles experiments. These test results will be discussed in the successive section. The coefficients of the controllers in this tests were as follows:
\begin{table}[h!]
  \begin{center}
    \caption{Controllers Coefficients}
    \label{tab:table1}
    \begin{tabular}{l|c|r} 
      \textbf{Controller} & $\mathbf{k_p}$ & $\mathbf{k_r}$ \\
      \hline
     Foot Length with Force Sensor & 0.00005 & 1\\
     \hline
      Foot Length with Bump Sensor & 0.005 & 1\\
      \hline
      Ankle Orientation & 0.0158
 & 6\\
    \end{tabular}
  \end{center}
\end{table}

\begin{figure}[]
\centering
\includegraphics[scale=0.5, trim ={3.0cm 8.0cm 2.0cm 7.0cm},clip]{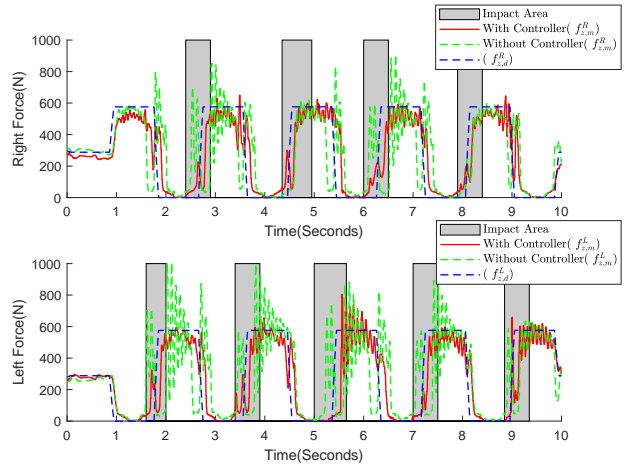}
\caption{Vertical forces applied to the soles of the robot's feet in the
presence and absence of the bump sensor controller layer}
      \label{fig:Verticalforces}
\end{figure}

\begin{figure}[]
\centering
\includegraphics[scale=0.6, trim ={4.0cm 8.5cm 4.0cm 8.0cm},clip]{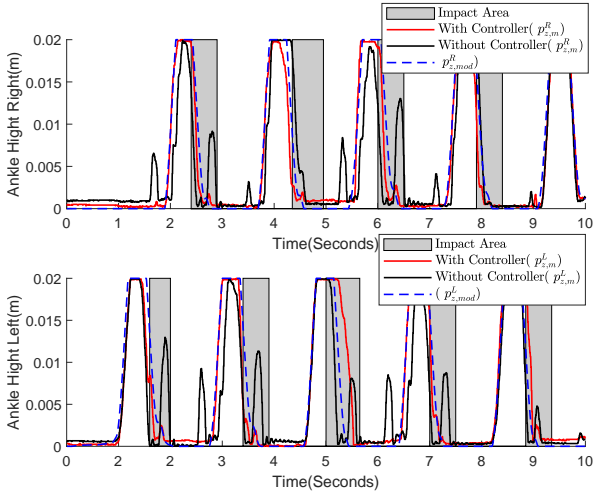}
\caption{Vertical distance of the robot's feet from the ground in the
presence and absence of the bump sensor controller layer}
      \label{fig:Verticaldistance}
\end{figure}
\subsection{Walking at the speed of  $1\frac{km}{h}$ }
As the speed of the robot increases, the destructive effects of the touchdown impact exacerbates the robot instability. The sensors presented in this article can help attenuate the effects of these impacts. In this test, the lower limb of the Surena V robot walked with a step length of 25 cm and a step time of 0.9 seconds (which is equivalent to moving at 1 km/h). In one of these tests, the controller that works with the feedback of the bump sensor was turned off, and in the other, the controller was turned on. 
Figure 7 shows the vertical forces measured on the soles of the robot in both tests. Figure 8 also shows the distance of the robot's feet from the ground measured by the bump sensor. The impact area in these figures is the end part of the swing phase and when the robot's foot hits the ground. As can be seen, when the bump sensor controller is off, the forces applied to the robot's feet increase. These forces appear as a form of impact at the end of the single support phase, which can be understood from the distance of the robot's feet from the ground in Figure 8. These impacts cause vibration in the upper body of the robot and make it challenging to maintain the robot’s balance. Moreover, increasing $k_p$ in (2) does not solve this problem, and if it exceeds a limit, it causes vibration in the movement of the robot. In the second test, where the bump sensor controller is used, these impacts have been significantly reduced, and the robot ankles have also traveled the modified trajectory well.
\subsection{Walking on inclined obstacles}
Our robot does not use a visual sensor such as  camera or LiDAR to perceive the environment. Therefore, if there is an obstacle on its way that is not considered in the trajectory planning, it will cause impacts during movement and loss of robot balance. Also, if this obstacle is inclined, the robot must adapt its foot orientation. In this experiment, the robot moves on two inclined obstacles that are not aware of them. The height of these obstacles is a maximum of 2.5 cm, with a slope of 7 to 12 degrees. Figure 1 shows the walking of the robot over these obstacles.
The ankle orientation controller plays the main role in adapting the robot's foot to the ground. Figure 9 and Figure 10  show the output of this controller, and the angle is given to the robot while crossing these obstacles. The control output when passing over these obstacles is more than at other times, which causes the robot's feet to adapt.

The most critical weakness of this controller is that if the obstacle does not collide with the sensors, the controller does not work well, such as moving the robot on rocks. We plan to solve this problem in future works.

\begin{figure}[]
\centering
\includegraphics[scale=0.4, trim ={0.0cm 13.0cm 0.0cm 0.0cm},clip]{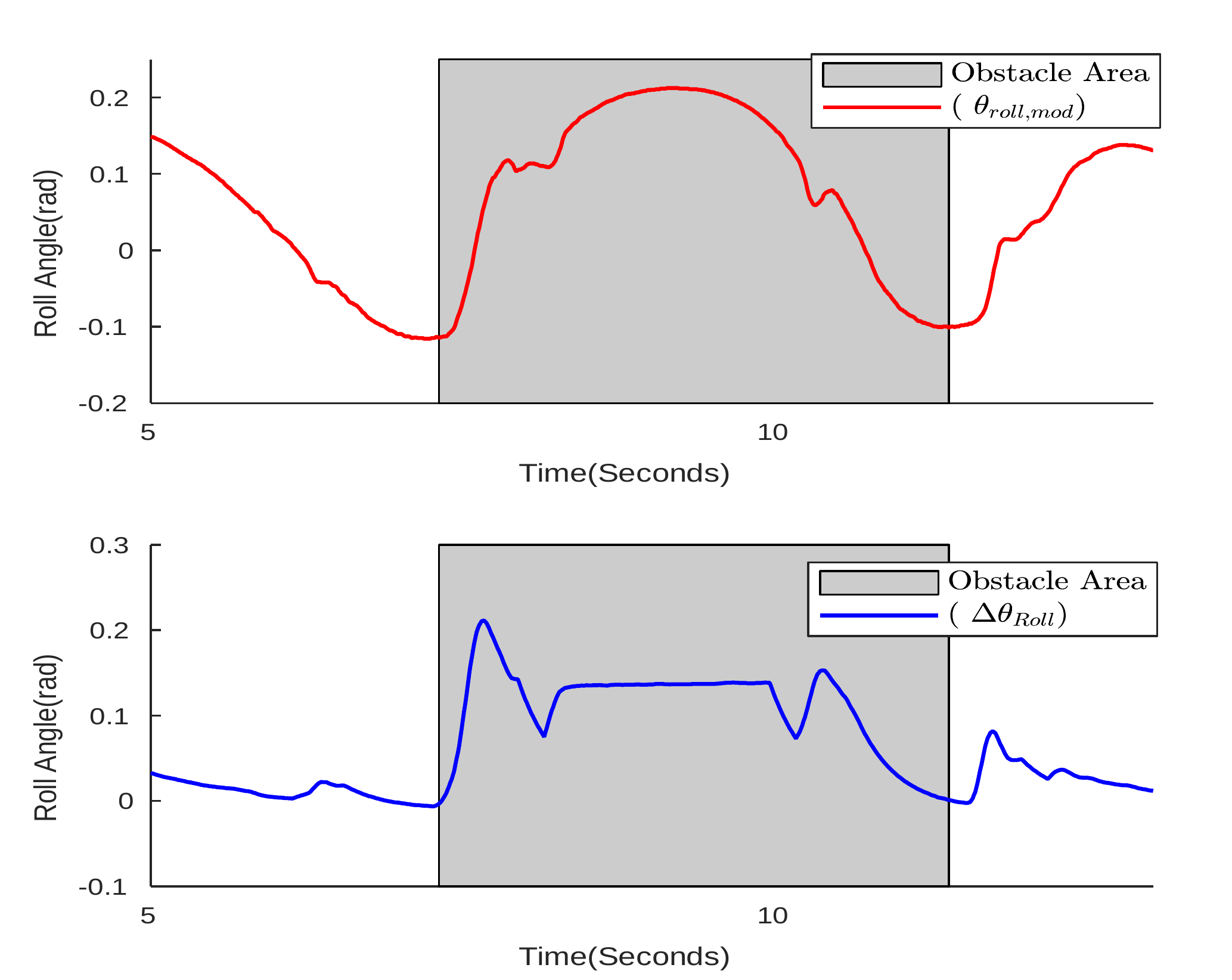}
\caption{The angle given to the robot's foot and the output of the
controller while crossing the rolled obstacle}
      \label{fig:anglerolled}
\end{figure}

\begin{figure}[]
\centering
\includegraphics[scale=0.4, trim ={0.0cm 11.0cm 0.0cm 0.0cm},clip]{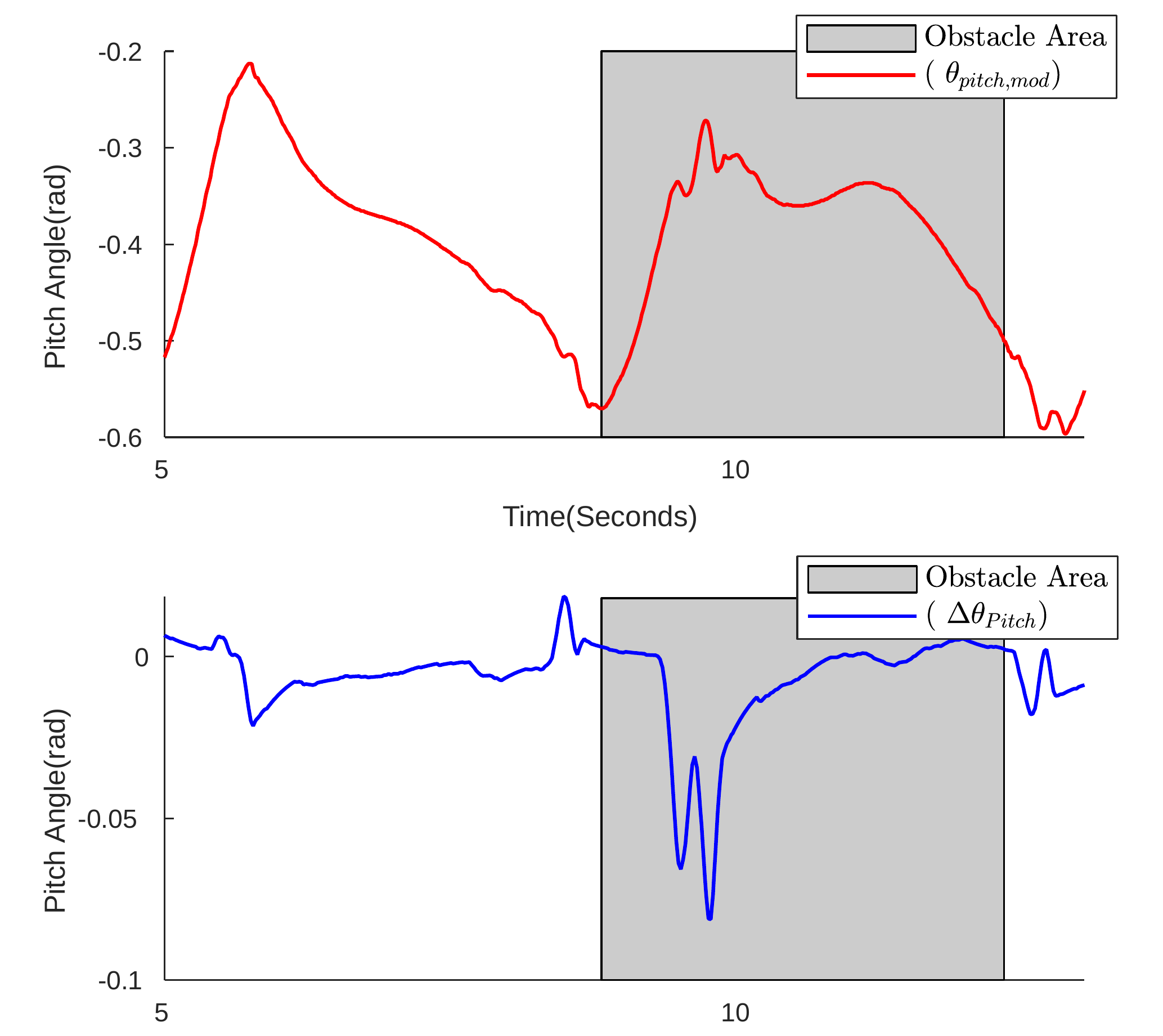}
\caption{The angle given to the robot's foot and the output of the
controller while crossing the pitched obstacle}
      \label{fig:anglepitched}
\end{figure}
\section{Conclusion}



In this paper, we presented a control framework and novel foot sensor for online foot adaption during locomotion on inclined obstacles and reducing the impact with the ground during swing to stance transition. We showed that robot is able to step on inclined obstacles without changing  the desired velocity. Considering that these sensors detect the distance to the ground before touching the ground, the controller drastically reduces the impact force, resulting an smooth swing-to-stance transition. Therefore, in addition to stepping on obstacles, it also compensates for the impact with the ground as a result of accumulative error due to the deflection of the links and connections. In future works, we are interested to use these sensors in combination with an admittance controller with torque feedback so that the robot can move on intensive rough terrains. Moreover, proposed architecture reduces the vibration of the robot and increases the accuracy of estimating the states of the robot. Using this architecture in conjunction with a push recovery controller\cite{shafiee2017push,shafiee2017robust,shafiee2019online}, we aim to perform locomotion on uneven terrain in the presence of significant external disturbances.
\addtolength{\textheight}{-8.7cm}   


\bibliography{bibliography}
\bibliographystyle{IEEEtran}

\end{document}